\documentclass[letterpaper, 10 pt, conference]{ieeeconf}  % Comment this line out
\IEEEoverridecommandlockouts                              % This command is only
\usepackage[utf8]{inputenc}
\usepackage[T1]{fontenc}

\usepackage{graphicx} % for pdf, bitmapped graphics files
\usepackage{epsfig} % for postscript graphics files
\usepackage{mathptmx} % assumes new font selection scheme installed
\usepackage{mathptmx} % assumes new font selection scheme installed
\usepackage{amsmath} % assumes amsmath package installed
\usepackage{amssymb}  % assumes amsmath package installed
\usepackage[dvipsnames]{xcolor}
\usepackage{float}
\usepackage[textsize=footnotesize,textwidth=1\linewidth]{todonotes}
\usepackage{multirow}
\usepackage[skip=2pt,font=small]{caption}
\usepackage{xcolor}
\usepackage{array}

\usepackage{booktabs}% More proffesional look of tables.
\usepackage{siunitx}% An awesome package for typesetting and manipulation numbers and units.
\usepackage{lipsum}% Example text
\usepackage{array}
\usepackage{gensymb}
\usepackage{cite}
\usepackage{hyperref}

\title{\LARGE \bf
Modeling and Validation of Soft
Robotic Snake
Locomotion}

\author{Dimuthu~D.~Arachchige$^{1}$, Yue Chen$^{2}$, and  Isuru~S.~Godage$^{1}$% <-this % stops a space
\thanks{$\!\!\!\!\!\!\!\!\!\!\!^{1}$Dimuthu~D.~Arachchige and Isuru~S.~Godage are with School of Computing, DePaul University, Chicago, IL, 60604.\,\,{\tt\small darachch@depaul.edu}
\, $^{2}$Yue Chen is with the Department of Mechanical Engineering, University of Arkansas Fayetteville, AR, 72701.
\vspace{1.5mm}
\newline
$\!\!\!\!$This work supported in part by the National Science Foundation grants IIS-1718755 and and IIS-2008797 and UARK Chancellor’s Fund for Innovation and Collaboration.
\vspace{1.5mm}
\newline
This paper has been submitted to IEEE International Conference on Robotics and Automation 2021.
}
}

\begin{document}
\maketitle
\thispagestyle{empty}
\pagestyle{empty}

\begin{abstract}
Snakes are a remarkable evolutionary success story. Many snake-inspired robots have been proposed over the years. Soft robotic snakes (SRS) with their continuous and smooth bending capability better mimic their biological counterparts' unique characteristics. Prior SRSs are limited to planar operation with a limited number of planar gaits. We propose a novel SRS with spatial bending and investigate snake locomotion gaits beyond the capabilities of the state-of-the-art systems. We derive a complete floating-base kinematic model of the robot and use the model to derive jointspace trajectories for serpentine and inward/outward rolling locomotion gaits. The locomotion gaits for the proposed SRS are experimentally validated under varying frequency and amplitude of gait cycles. The results qualitatively and quantitatively validate the SRS ability to leverage spatial bending to achieve locomotion gaits not possible with current SRS. 
\end{abstract}

\section{Introduction\label{sec:Introduction}}
Snakes are highly capable animals with a wide range of habitats, including hostile deserts, dense tropical forests, and uninhabitable marshes. One key feature that makes snakes unique in their ability to navigate in different terrains using various locomotion gaits is supported by their long, high degrees of freedom (DoF) slender bodies. The body's continuous and smooth bending structure enables snakes to overcome numerous environmental challenges such as climbing, swimming despite having no sophisticated appendages such as limbs. Their high DoF body can generate a range of locomotion gaits such as lateral undulation, rectilinear movement, sidewinding, concertina movement. Snake-like robots can harness this traversability in different and challenging terrains in applications such as inspection tasks, reconnaissance, search, and rescue operations. Their small cross-section to length ratio allows them to gain access through tight spaces, little narrow openings (i.e., sewage lines), and perform assigned operations. Robotics have developed various snake robot prototypes \cite{hirose1993biologically,hirose2004biologically,hirose2009snake,wright2007design,wright2012design,transeth20083,crespi2008online,luo2014design, onal2013autonomous,qin2018design} to meet the aforementioned application challenges.

\begin{figure}[t] 
	\centering
	\includegraphics[width=1\linewidth]{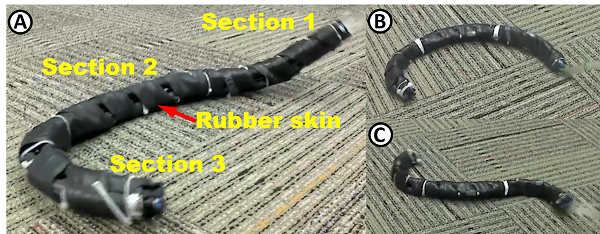}
	\caption{(A) The proposed Soft robotic snake (SRS) prototype, (B) Replicating rolling locomotion, and (C) Replicating serpentine locomotion.}
	\label{fig:IntrodudctoryImage} 
\end{figure}

\subsection{Prior Work\label{subsec:Prior-Work}}

Prior work on snake robots mainly evolved with modular rigid robots \cite{hirose1993biologically,hirose2004biologically, wright2012design,transeth20083,crespi2008online}. Rigid robotic snakes (RRS) use jointed rigid links to achieve bending motion and generate locomotion. RRS, with few rigid links, limits the robot from achieving the smooth bending observed in snakes and, therefore, can affect locomotion. On the other hand, soft robotic snakes (SRS), constructed mainly from fluidic muscle actuator-powered bodies, can better leverage smooth bending to imitate snake body movements. However, 
%to date, 
the latest SRS prototypes are limited to planar bending deformation, limiting the number of gaits they can demonstrate. In addition, snakes use differential friction property of their skin to generate locomotion in planar gaits such as serpentine gait. Noting the absence of such quality in current SRS, it is extremely challenging to propel robots solely via planar locomotion gaits. Consequently, SRSs rely on wheels to circumvent uniform friction and generate locomotion \cite{luo2018orisnake,qin2018design}. Thus, we posit that it is essential to exploit the out-of-plane deformation (spatial bending) to replicate locomotion gaits such as rolling and sidewinding. The SRS reported in \cite{onal2013autonomous, luo2014theoretical} generates serpentine locomotion but utilizes wheeled bases to generate friction anisotropy necessary for movement. The `WPI SRS' \cite{luo2015refined} reported 10-fold locomotion speed increase than the previous version \cite{luo2014design}. The SRS presented in \cite{luo2015slithering} and \cite{luo2017toward} is self-contained with integrated sensing and feedback control. It improves the accuracy of dynamic undulatory locomotion.  We propose a novel SRS with spatial bending capabilities to address these limitations.

\subsection{Contribution\label{subsec:Contribution}}
The main contributions of this work are (a) propose a novel SRS with spatial bending capability, (b) derive complete floating base kinematic model, (c) derive jointspace trajectories of serpentine, inward rolling, and outward rolling snake locomotion gaits,  (d) experimentally validate the said locomotion gaits for a range of pressure-frequency combinations on the propose SRS, (e) demonstrate the need for spatial bending to overcome the limitations of friction anisotropy present in practical SRS. The experimental results show that SRS can successfully track the spatial shape trajectories for all the gaits. This is the first SRS to utilize spatial bending capability and demonstrate meaningful locomotion (using inward/outward rolling gait) without wheels and utilizing the friction forces generated by skin-ground interactions.

%We introduce a novel multi-DOF SRS which can replicate snake locomotion gaits in high fidelity. The state-of-the-art soft robotic snakes show limitations on replicating locomotion gaits, which require out-of-plane deformation. Most of the available soft robotic snakes only focus on validating planer gaits, more commonly serpentine motion. The systems capable of replicating out-of-plane deformation are designed with added complications such as wheels, making the robot design a complex system on its own. The literature shows that the robot cannot exploit the essential third dimension without such a three-dimensional support system. We introduce this new multi-section snake robot, which can exploit the third dimension within its structure without added complications. We achieve it with the help of spatial bending. In the field of soft robotics, this is the first time a robotic snake can exploit the out of plane deformation through its skin without added complications.
%\todo[inline]{re-write contribution section. right now its just the same arguments made against current SRS, not the specific contributions of our work}

\section{System Model\label{sec:systemModel}}
\subsection{Prototype Description\label{sub:prototypeDescription}}

\begin{figure}[tb] 
	\centering
	\includegraphics[width=1\linewidth,height=.7\linewidth]{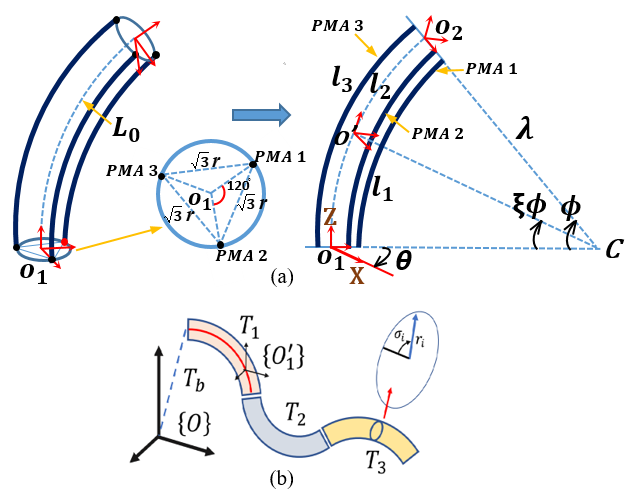}
	\caption{ A schematic of the SRS and a schematic of a single section.}
	\label{fig:schematicKinematicModel} 
\end{figure}
Fig. \ref{fig:IntrodudctoryImage} shows the prototype of the proposed SRS. It is made of three serially attached soft bending modules (or sections). They are powered by McKibben type extending mode pneumatic muscle actuators (PMA). These actuators proportionally extend the body to supplied pneumatic pressure up to 4~bars. In a single soft module, 3 PMAs are assembled at $\frac{\pi }{3}$ separation from each other to ensure symmetric spatial bending and facilitate room to route pneumatic supply lines within the module for a streamlined physical profile required for a slender snake-like body. Each PMA has an unactuated length, $L_{i0}$=0.15~m, and can extend by 0.065~m at 4~bar pressure. Rigid 3D printed mounting frames (made of ABS thermoplastic) are used to mount PMAs at $r_{i}$=0.0125~m from the centerline of soft modules (Fig. \ref{fig:IntrodudctoryImage}). Similarly, laser-cut plastic constrainer plates of $r_{i}$ and 0.0025~m thickness are used along the soft modules' length to maintain PMAs parallel to soft modules' central axis with a $r_{i}$ clearance from the central axis. The constrainer plates also provide structural strength for this long and slender SRS to maintain its structural integrity during locomotion and generate reaction forces required for locomotion. All actuators are bundled within the soft module as a single unit, similar to the continuum sections reported in \cite{godage2016dynamics}. The pressure differential among PMAs of a soft module causes the module to bend in any direction or extend axially. Thus we can control the bends synchronously in order to generate various types of robot locomotion gaits. Soft modules are then connected via the mounting frames at a $\frac{\pi}{3}$ offset to each other to create the SRS (Fig. \ref{fig:IntrodudctoryImage}). Finally, the SRS is wrapped with rubber skin, and without the pneumatic supply lines, it weighs close to 0.3~kg.

\subsection{Kinematic Model\label{sub:kinematicModel}}

Consider the schematic of any $i^{th}$ module ($i \in \{1,2,3\}$) of the SRS, as shown in Fig. \ref{fig:schematicKinematicModel}-a. It depicts three mechanically identical variable length actuators (PMAs) with an unactuated length $L_{i0}\in\mathbb{R}$ and length change $l_{ij}\left(t\right)\in\mathbb{R}$, where $j\in\left\{ 1,2,3\right\} $ is the actuator index, and $t$ is the time. Hence, the length of an actuator at any time is $L_{ij}=L_{i0}+l_{ij}(t)$. The kinematic model of the proposed SRS can be formulated by extending the modal kinematics proposed by Godage et al.  in \cite{godage2015modal}. Let the joint space vector of any $i^{th}$ soft robot module be $\boldsymbol{q}_{i}=\left[l_{i1}\left(t\right),\,l_{i2},\,l_{i3}\left(t\right)\right]^{T}$. Utilizing the results from \cite{godage2015modal}, we can derive the homogeneous transformation matrix (HTM) at any point along the neutral axis of the $i^{th}$ soft  module, $\mathbf{T}_{i}\in\mathbb{SE}^{3}$, as

\vspace{-5mm}
\begin{align}
	\mathbf{T}_{i}\left(\boldsymbol{q_{i}},\xi_{i}\right) & =\left[\!\!\!\begin{array}{cc}
		\mathbf{R}_{i}\left(q_{i},\xi_{i}\right) & \!\!\!\boldsymbol{p}_{i}\left(q_{i},\xi_{i}\right)\\
		\mathit{\boldsymbol{0}} & 1
	\end{array}\!\!\!\right]\left[\!\!\!\begin{array}{cc}
		\mathbf{R}_{z}\left(\sigma_{i}\right) & \!\!\!\mathcal{\mathit{\boldsymbol{p}_{x}\left(\delta_{i}\right)}}\\
		\mathit{\boldsymbol{0}} & 1
	\end{array}\!\!\!\right]\label{eq:ith_kin}
\end{align}
where $\mathbf{R}_{i}\in\mathbb{SO}^{3}$ is the rotational matrix, $\boldsymbol{p}_{i}\!\in\mathbb{R}^{3}$ is the position vector $\xi_{i}\in\left[0,1\right]$ is a scalar to define points along the soft module with 0, and 1 denotes the base and the tip, and $I_3$ is the rank 3 identity matrix. In addition to the previous results in \cite{godage2015modal}, note that we introduce two HTMs with $\mathbf{R}_{z}\in\mathbb{SO}^{3}$ is the rotation matrix about the $+Z$ axis, and $\boldsymbol{p}_{x}\in\mathbb{R}^{3}$ is the position offset along the $+Z$ of $ O_{i}^{\prime}$ where $\sigma_{i}\in\left[0,2\pi\right]$. 

Utilizing \eqref{eq:ith_kin} with a floating coordinate system, $\mathbf{T}_{b}\in\mathbb{SE}^{3}$, the complete kinematic model along the body of the snake robot (Fig. \ref{fig:schematicKinematicModel}-b) is given by

\vspace{-5mm}
\begin{align}
	\mathbf{T}\left(\boldsymbol{q}_{b},\boldsymbol{q},\boldsymbol{\xi}\right) & =\mathbf{T}_{b}\left(\boldsymbol{q}_{b}\right)\prod_{i=1}^{3}\mathbf{T}_{i}\left(\boldsymbol{q_{i}},\xi_{i}\right)\nonumber \\
	& =\left[\begin{array}{cc}
		\mathbf{R}\left(\boldsymbol{q}_{b},\boldsymbol{q},\xi\right) & \boldsymbol{p}\left(\boldsymbol{q}_{b},\boldsymbol{q},\xi\right)\\
		\mathit{\boldsymbol{0}} & 1
	\end{array}\right]\label{eq:complete_kin}
\end{align}
where $\boldsymbol{q}_{b}=\left[x_{b},y_{b},z_{b},\alpha,\beta,\gamma\right]$ are the parameters of the floating coordinate system with $\left[x_{b},y_{b},z_{b}\right]$ denote the translation and $\left[\alpha,\beta,\gamma\right]$ denote the XYZ Euler angle offset of the base of 
%the 
module 1, i.e., the origin of the robot coordinate frame, with respect to $\left\{ O\right\} $. The composite vector $\boldsymbol{q}=[\boldsymbol{q}_{1},\boldsymbol{q}_{2,},\boldsymbol{q}_{3}]\in\mathbb{R}^{9}$ and $\boldsymbol{\xi}=\left[0,3\right]\in\mathbb{R}$.

\section{Trajectory Generation\label{sec:TrajectoryGeneration}}

\begin{figure}[tb]
	\centering
	\includegraphics[width=1\linewidth,height=.4\linewidth]{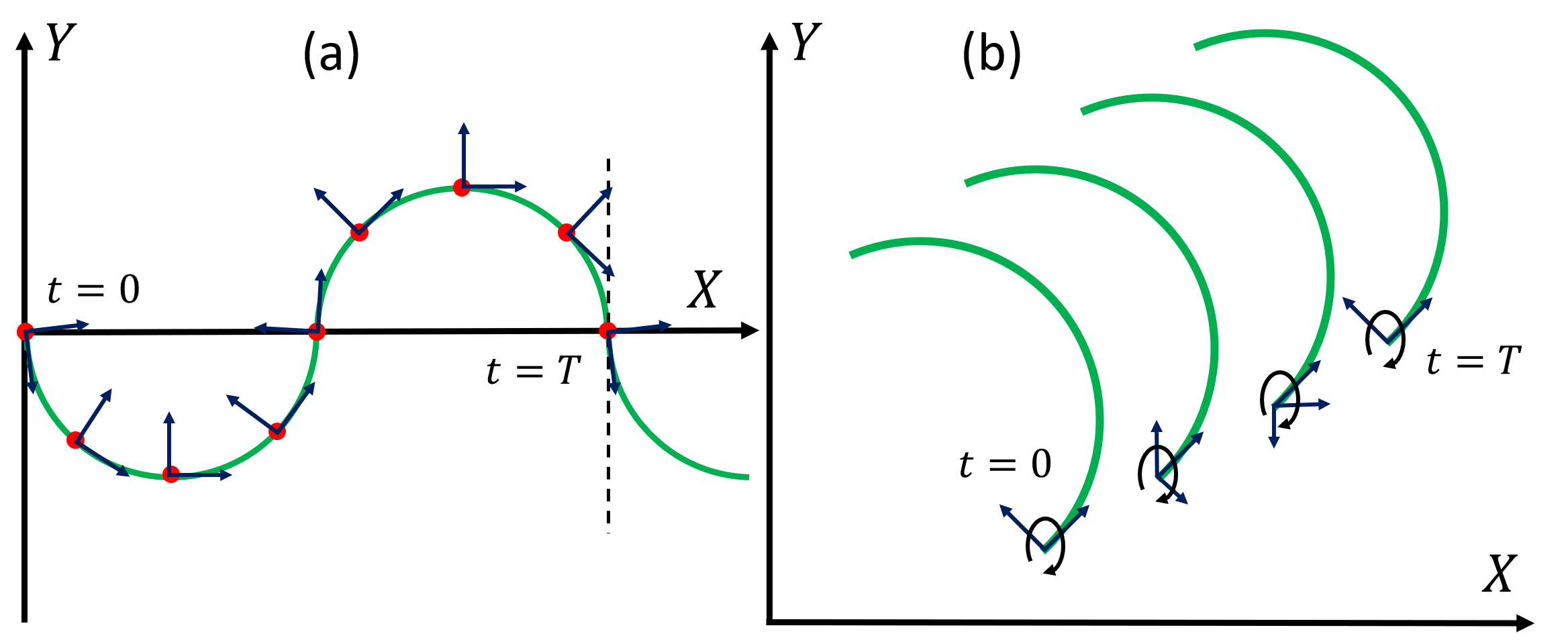}
	\caption{Trajectory sampling at different time instances within a gait cycle for (a) serpentine gait and (b) outward rolling gait.}
	\label{fig:trajsampling}
\end{figure}

In this work, we consider serpentine, inward rolling, and outward rolling locomotion gaits, which are cyclic in nature. We can mathematically model these gaits associated with different locomotion gaits to generate the taskspace trajectories thereof. Fig. \ref{fig:trajsampling} shows the taskspace trajectories of serpentine and inward rolling locomotion. The serpentine curve is given by

\vspace{-5mm}
\begin{align}
	x\left(s\right) & =\int_{0}^{s}\cos\left(a\cos\left(b\sigma\right)+c\sigma\right)d\sigma\nonumber \\
	y\left(s\right) & =\int_{0}^{s}\sin\left(a\cos\left(b\sigma\right)+c\sigma\right)d\sigma
	%z\left(s\right) & =0\nonumber 
	\label{eq:serpentineEq}
\end{align}
where $a=-\frac{pi}{4}$, $b=\frac{pi}{4}$, $c=0$, and $z\left(s\right)=0$.

\begin{figure}[tb] 
	\centering
	\includegraphics[width=1\linewidth]{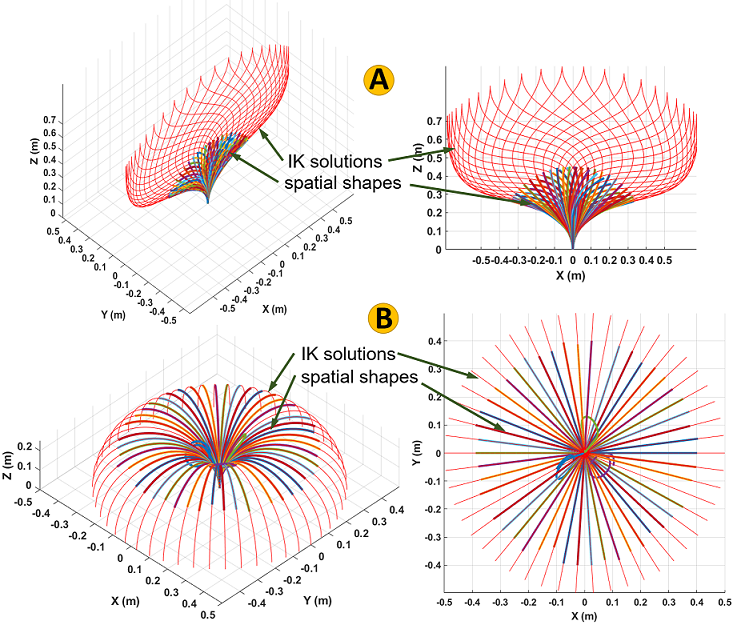}
	\caption{Trajectory curves in the robot coordinate frame: (A) Serpentine locomotion, (B) Rolling locomotion.}
	\label{fig:TotalMathematicalCurves} 
\end{figure}

The rolling gait can be modeled as a mathematical curve -- in this case, the displacement of a circular arc with radius $r_R$, in the rolling gait. One cycle of the rolling gait is defined as the rotation of the curve about its longitudinal axis (Z-axis in the robot coordinate frame), and the robot displacement on the X-Y plane is due to the thickness of the robot that can be derived as $2\pi r_i$, where $r_i$ is the radii of soft modules. From one period of a gait, we derive spatial shapes from the curve at different time intervals within one locomotion cycle to obtain different spatial shapes resembling the snake shape at each of those instances. Similarly, Fig. \ref{fig:trajsampling} shows the progression of the serpentine gait and inward rolling gait in one cycle. The dots along the serpentine curve, shown in Fig. \ref{fig:trajsampling}-a denote the robot origin of the robot coordinate frame at the sampling time instances. Similarly, Fig. \ref{fig:trajsampling}-b shows the rolling locomotion progression at different time instances. Next, we map these spatial shape trajectories to jointspace trajectories of the SRS. 
The kinematic model we derived in Sec. \ref{sub:kinematicModel}, given by \eqref{eq:complete_kin} has a floating base coordinate that denotes the origin of the SRS with respect to the inertial frame. However, the floating DoFs are redundant for generating the bending shape of the robot. We follow the steps outlined below to identify the shapes and project those shapes to the SRS's coordinate system. 

From the progression of the locomotion gait at different times within a period, we define the curves' origin at those points (Fig. \ref{fig:trajsampling}). Without losing generality, consider any $n^{th}$ point along the curve.  We then derive a local coordinate frame with respect to the inertial frame at those points. For instance, Fig. \ref{fig:trajsampling}-a shows how we derive coordinate frames at locations shown as dots. For the serpentine curve, a tangential line to the curve and a line normal to this tangential line is defined as the unit vectors along the local Y and X axes as $\overrightarrow{e}_{X}$  and $\overrightarrow{e}_{Y}$ respectively. The unit vector of the Z-axis is derived from $\overrightarrow{e}_{X} \times \overrightarrow{e}_{Y}$. This information and the position vector of the location under consideration are then used to derive the HTM at that point with respect to the inertial frame. In the case of the rolling gait, due to the curve's rotation about its longitudinal axis, the robot coordinate frame is 3D (Fig. \ref{fig:trajsampling}-b).

Utilizing these HTMs at each instance, we project the related mathematical curve on to the body coordinate frame. This step is repeated for all the subsequent taskspace curve shapes within a gait cycle. Figs. \ref{fig:TotalMathematicalCurves}-A and B show the spatial shapes for serpentine and rolling motions projected on to the robot coordinate frame, respectively. Next, we derive the jointspace variables for each spatial shape, such that map the SRS shape to the curves under consideration. To that end, we formulate the shape matching as an optimization problem. Utilizing the kinematic model given in \eqref{eq:complete_kin}, we define 31 uniformly distributed points (10 points per soft module) along the SRS neutral axis by sampling $\xi$. Then we define the cost function given by

\vspace{-5mm}
\begin{align}
f_{cost} & =\sum_{k=1}^{31}\left\Vert \boldsymbol{p}\left(\boldsymbol{0},\boldsymbol{q},\xi_{k}\right)-f_{gait}\left(s\right)\right\Vert +\sum_{i,j\in\left\{ 1,2,3\right\} }l_{ij}^{2}\label{eq:costfun}
\end{align}
where $0\leq \xi_k\leq 3$ and $f_{gait}$ is the locomotion gait shape to which the SRS is fitted, and $s$ defines the points along $f_{gait}$. 

\begin{figure}[tb] 
	\centering
	\includegraphics[width=1\linewidth]{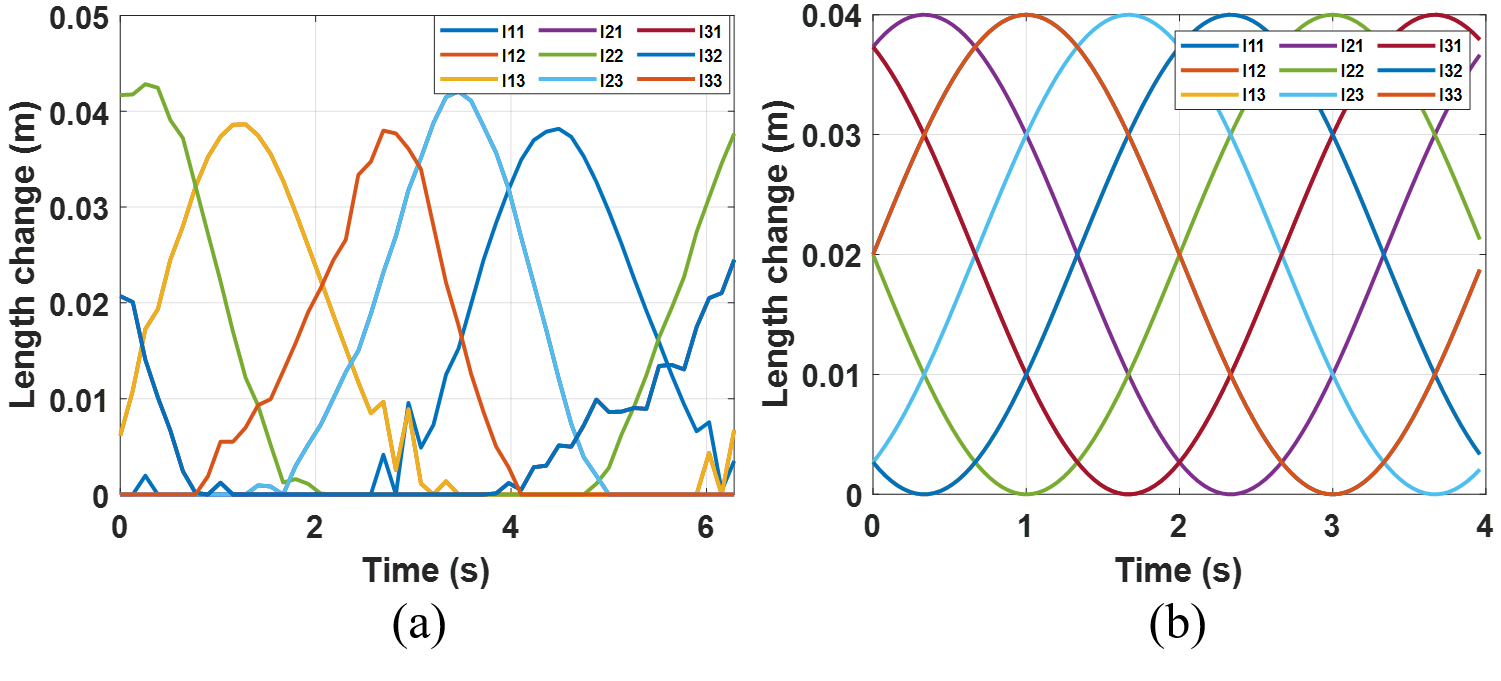}
	\caption{Jointspace trajectories of (a) serpentine gait, (b) rolling gait. The serpentine gait's nonsmooth trajectories are because, unlike rolling gait, of the imperfect matching of optimal SRS shapes do gait taskspace shapes.}
	\label{fig:LengthTrajectories} 
\end{figure}

The SRS concerned here are constructed from three extensible soft modules, and the latter term of \eqref{eq:costfun} ensures that the optimization solution has the least total extension. This ensures the jointspace trajectories to stay smooth without unnecessary length changes between solutions. We use Matlab's global constrained optimization routine to optimize the SRS shape and save the jointspace solution. Fig. \ref{fig:TotalMathematicalCurves} shows the matched SRS shapes in thick lines. There are instances where the optimization failed. However, this has little effect on the overall jointspace trajectory due to the dense sampling of gait cycles. One may rerun the optimization routine with different initial conditions to obtain a solution. We repeat the process for all the spatial mathematical curves associated with each instance of the locomotion gaits. Fig. \ref{fig:LengthTrajectories} shows the jointspace trajectories for serpentine and rolling snake locomotion gait for several cycles. Note that the outward rolling trajectory is the reverse of the inward rolling trajectory. The cycle frequency of these gaits can be scaled by adjusting the cycle period. Similarly, the change of amplitude of the jointspace values results in more bending. In the case of rolling locomotion, this means a reduced $r_R$, i.e., rolling arc radius, whereas it increases the serpentine gait's amplitude.

%\vspace{-5mm}

\section{Experimental Validation\label{sec:ExperimentalValidation}}

\subsection{Experimental Setup}
\begin{figure}[tb] 
	\centering
	\includegraphics[width=1\linewidth,height=.6\linewidth]{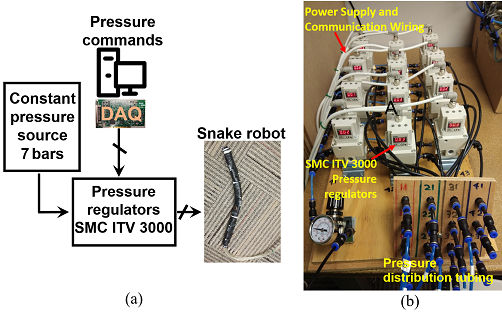}
	\caption{(a) Robot actuation setup, (b) Pressure regulator assembly.}
	\label{fig:operationdiagram_pressurevalvesetup} 
\end{figure}

Fig. \ref{fig:operationdiagram_pressurevalvesetup}a shows the overall experimental setup of the snake robot. It consists of an input pressure source,  pressure controllers, pressure command interface, and the SRS. The setup's input pressure is a constant 8~bar pressure supplied by an air compressor. The pressure to the PMAs is controlled by SMC ITV3000 series digital proportional pressure regulators and supplied via 4~mm diameter tubes. The commands to the pressure regulators are generated through a Matlab Simulink Desktop Realtime model and interfaced via a 0-10~V voltage signal using a National Instrument DAQ card. The SRS is tested on a carpeted floor, as shown in Fig. \ref{fig:RollingSerpentineimage}.

\subsection{Testing Procedure\label{subsec:testing}}

The SRS prototype is tested for three snake locomotion gaits; serpentine, inward rolling, and outward rolling presented in Sec. \ref{sec:TrajectoryGeneration}. The jointspace trajectories given in Fig. \ref{fig:TotalMathematicalCurves} depicts the length changes of PMAs. As the control the pressure to PMAs, we first need to establish a mapping between the PMA pressure input and the resulting length changes. We note that the SRS needs to work with relatively fast movements to generate highly dynamic locomotion trajectories. Therefore, an empirical and static length-to-pressure mapping-based approach may result in low accuracy. We examined the PMA length changes under for 1~Hz rectified triangular wave and found that $P_{ij}=100l_{ij}$ provides acceptable results where $P_{ij}$ is the pressure in bars to control the joint variable $l_{ij}$ (i.e., length change). We tested the gait trajectories for 15~s in pressure amplitudes \{1,2,3,4\}~bars and frequencies \{0.25,0.5,0.75,1.0\}~Hz combinations, totaling 16 experiments per gait type. The SRS motion is captured using a fixed camera station. We applied image perspective projection \cite{ohta1981obtaining} to track the robot movements.

\subsection{Serpentine Motion\label{subsec:SerpentineMotion}}

\begin{figure*}[tb] 
	\centering
	\includegraphics[width=1\textwidth]{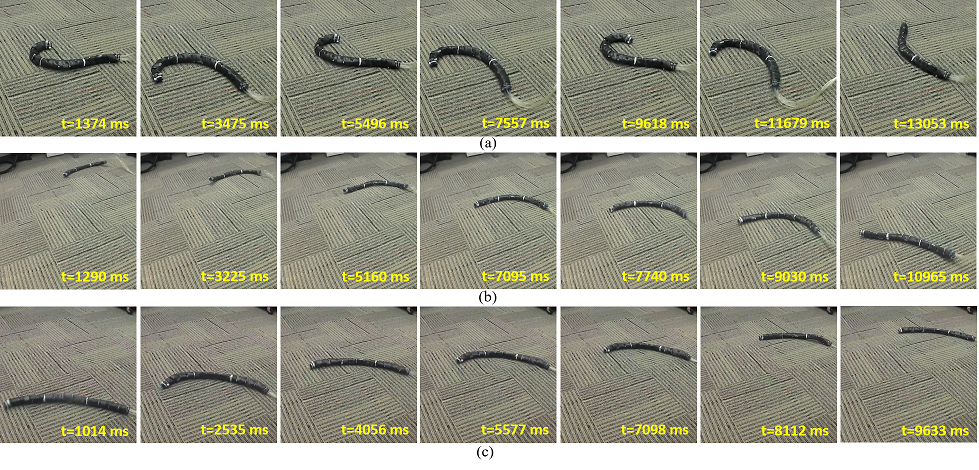}
	\caption{(a) Serpentine gait at 4~bar-0.25~Hz, (b) Inward rolling and (c) Outward rolling gaits at 4~bar-1.00~Hz pressure-frequency combinations.}
	\label{fig:RollingSerpentineimage} 
\end{figure*}

Fig. \ref{fig:RollingSerpentineimage}-a shows a chronological shape change of the robot for the serpentine motion at 4~bar pressure amplitude and 0.25~Hz frequency. We found that this combination replicates the best serpentine patterns. Fig. \ref{fig:MovementTracking}:A-C show how the proposed SRS behaves during serpentine motion at 2~bar-0.25~Hz, 4~bar-0.25 Hz, and 4 bar-1.00 Hz, pressure-frequency combinations. Refer to the accompanying multimedia submission for the videos of the experiments. Further, we did not observe any meaningful serpentine locomotion in the axial direction, as shown in Fig. \ref{fig:MovementTracking}:A-C. Further, it is discovered from the numbers indicated in Table \ref{Table:travellingdisplacementrolling} under the traveling velocity of the serpentine gait. Here, the robot wobbles around without making any progress. This is a common problem with snake robots. It is mainly due to insufficient friction difference in the robot skin in normal and tangential directions. In the real world, snakes have different friction coefficients in these directions. In our robot, the skin is made of a rubber surface, resulting in the same friction coefficient in all directions. Therefore the robot cannot generate a forward propagation force in the axial direction. As expected, it does not result in any forward locomotion with serpentine patterns. This is one of the main reasons why the prior work involving planer SRS \cite{onal2013autonomous,luo2014theoretical,luo2014design,luo2015refined,luo2015slithering,luo2017toward,qin2018design,luo2018orisnake} uses wheels. The wheels generate low friction in the axial direction and high friction in the normal direction.

\subsection{Inward Rolling Motion\label{subsec:InwardRollingMotion}}
 
Fig. \ref{fig:RollingSerpentineimage}-b shows a chronological shape change of the robot during inward rolling at 4~bar actuation pressure and 1.0~Hz frequency, which stands as the best combination to illustrate inward rolling locomotion at the highest velocity. Figs. \ref{fig:MovementTracking}:D-F show how the robot behaves during inward rolling at 2~bar-0.25~Hz, 3 bar-0.50 Hz, and 4 bar-1.00 Hz,  pressure-frequency combinations. We observed that the robot could successfully replicate inward rolling locomotion at all pressure and frequency combinations except at a pressure as low as 1~bar. At 2~bar,  when the moving frequency was high as 1~Hz, the robot occasionally flipped back in the opposite direction. Further, the robot shows a low bending curvature at higher actuation pressure and frequency combinations. This is mainly due to air pressure not reaching the PMAs in realtime through long pneumatic lines. In both inward and outward rolling gaits, the highest traveling velocity is observed when the applied pressure-frequency combination is at its maximum value, i.e., 4~bar-1.00~Hz (Fig. \ref{fig:MovementTracking}:F-I and Table \ref{Table:travellingdisplacementrolling}). The SRS's rolling locomotion exploits 3D bending to rotate about its longitudinal axis, and this is the first time rolling locomotion is achieved by a SRS. Unlike serpentine gait, we demonstrate that SRS locomote successfully via rolling gaits. Here, the friction is applied in the rolling direction and supports rolling without any interference, similar to moving a continuous wheel. We observed similar performance in rolling in the reverse direction (Sec. \ref{subsec:OutwardRollingMotion}). 

\begin{table}[tb]
	\setlength{\tabcolsep}{5pt} %% default is 6pt
	\centering
	\caption{Traveling velocity of robot locomotion gaits. Velocity is computed by dividing the mean of the X and Y displacements by the experiment time.}
	\label{Table:travellingdisplacementrolling}
	\begin{tabular}{|c|c|c|c|c|c|c|c|} 
		\hline
		\multirow{3}{*}{\begin{tabular}[c]{@{}c@{}} \textbf{Pressure }\\\textbf{amplitude}\\(bar)\\ \end{tabular}} & \multirow{3}{*}{\begin{tabular}[c]{@{}c@{}}\textbf{Frequency}\\(Hz) \end{tabular}} & \multicolumn{6}{c|}{\begin{tabular}[c]{@{}c@{}}\textbf{Travelling velocity}\\(cm/s)\end{tabular}} \\ 
		\cline{3-8}
		&  & \multicolumn{2}{c|}{\begin{tabular}[c]{@{}c@{}}Serpentine\\motion \end{tabular}} & \multicolumn{2}{c|}{\begin{tabular}[c]{@{}c@{}}Inward~\\rolling \end{tabular}} & \multicolumn{2}{c|}{\begin{tabular}[c]{@{}c@{}} Outward~\\rolling\end{tabular}} \\ 
		\cline{3-8}
		&  & v\_x & v\_y & v\_x & v\_y & v\_x & v\_y \\ 
		\hline
		\multirow{4}{*}{1.0} & 0.25 & 0.00 & 0.00 & 0.00 & 0.00 & 0.00 & 0.00 \\ 
		\cline{2-8}
		& 0.50 & 0.00 & 0.00 & 0.00 & 0.00 & 0.00 & 0.00 \\ 
		\cline{2-8}
		& 0.75 & 0.00 & 0.00 & 0.00 & 0.00 & 0.00 & 0.00 \\ 
		\cline{2-8}
		& 1.00 & 0.00 & 0.00 & 0.00 & 0.00 & 0.00 & 0.00 \\ 
		\hline
		\multirow{4}{*}{2.0} & 0.25 & 0.07 & 0.13 & 1.14 & 2.21 & 0.44 & 2.20 \\ 
		\cline{2-8}
		& 0.50 & 0.12 & 0.16 & 2.11 & 3.21 & 1.29 & 2.02 \\ 
		\cline{2-8}
		& 0.75 & 0.14 & 0.21 & 2.98 & 3.99 & 2.15 & 1.82 \\ 
		\cline{2-8}
		& 1.00 & 0.08 & 0.14 & 3.43 & 4.32 & 3.01 & 1.48 \\ 
		\hline
		\multirow{4}{*}{3.0} & 0.25 & 0.28 & 0.43 & 3.99 & 4.87 & 3.33 & 2.08 \\ 
		\cline{2-8}
		& 0.50 & 0.49 & 0.65 & 4.05 & 5.72 & 5.69 & 0.12 \\ 
		\cline{2-8}
		& 0.75 & 0.59 & 0.71 & 5.16 & 5.89 & 4.21 & 2.19 \\ 
		\cline{2-8}
		& 1.00 & 0.28 & 0.55 & 5.23 & 7.89 & 4.89 & 1.91 \\ 
		\hline
		\multirow{4}{*}{4.0} & 0.25 & 0.76 & 0.36 & 3.12 & 7.55 & 4.03 & 3.91 \\ 
		\cline{2-8}
		& 0.50 & 0.63 & 0.28 & 4.16 & 9.11 & 5.79 & 2.78 \\ 
		\cline{2-8}
		& 0.75 & 0.21 & 0.27 & 4.98 & 9.67 & 6.39 & 4.55 \\ 
		\cline{2-8}
		& 1.00 & 0.09 & 0.18 & 5.61 & 10.11 & 6.56 & 6.75 \\
		\hline
	\end{tabular}
	\vspace{-5mm}
\end{table}

\subsection{Outward Rolling Motion\label{subsec:OutwardRollingMotion}}

\begin{figure*}[tb] 
	\centering
	\includegraphics[width=1\textwidth]{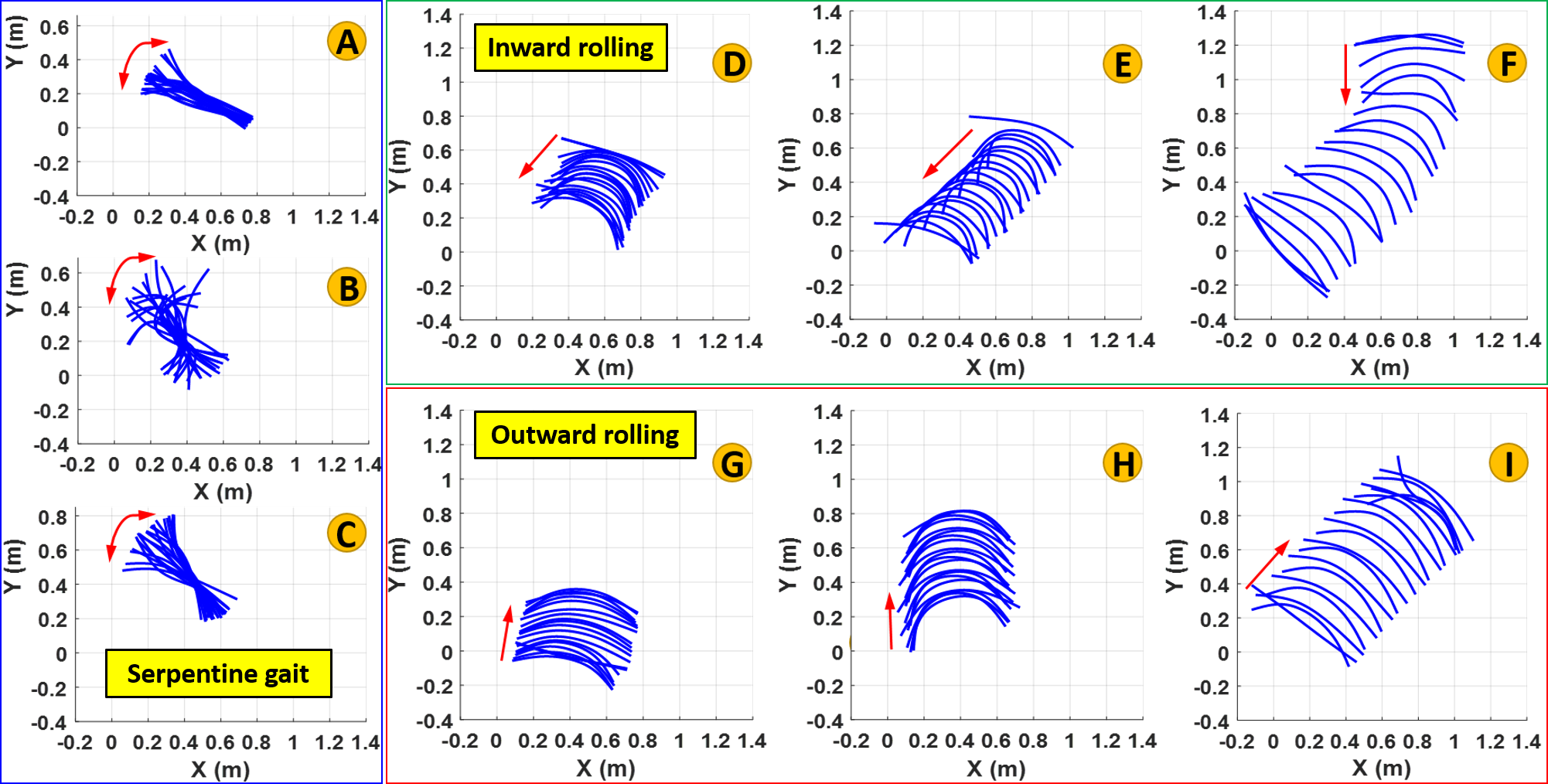}
    \caption{Tracking of robot movement for serpentine gait at (A) 2~bar-0.25~Hz, (B) 4~bar-0.25~Hz, and (C) 4~bar-1.00~Hz pressure-frequency combinations; Inward rolling gait at (D) 2~bar-0.25~Hz, (E) 3~bar-0.50~Hz, and (F) 4~bar-1.00~Hz pressure-frequency combinations; and Outward rolling gait at (G) 2~bar-0.25~Hz, (H) 3~bar-0.50~Hz, and (I) 4~bar-1.00~Hz pressure-frequency combinations.}
	\label{fig:MovementTracking} 
\end{figure*}

Fig. \ref{fig:RollingSerpentineimage}-c shows a chronological shape change of the robot during outward rolling motion at 4~bar actuation pressure and 1.0~Hz frequency. We found that this is the best combination to illustrate outward rolling locomotion at maximum velocity. Fig. \ref{fig:MovementTracking}:G-I show how the robot behaves during inward rolling at 2~bar-0.25~Hz, 3~bar-0.50~Hz, and 4~bar-1.00~Hz,  pressure-frequency combinations. Table \ref{Table:travellingdisplacementrolling} shows the calculated traveling velocity of all three locomotion gaits based on image tracking results in Fig. \ref{fig:MovementTracking}. Here, the robot performs outward rolling towards opposite to its curve opening. The robot cannot perform very well as much as in the other direction. The friction is applied opposite the rolling direction, and the friction force interferes with the generated rolling thrust. It is witnessed by relatively low velocities recorded in Table \ref{Table:travellingdisplacementrolling}.

\subsection{Discussion\label{subsec:discussion}}

The tracking results in Fig. \ref{fig:MovementTracking} show that the SRS replicates the three locomotion gaits very well. When the actuation pressure is low at 1 bar regardless of frequency or the locomotion gait type, the robot showed almost no movements. This is mainly due to the associated ~1.0-bar deadzone present in PMAs where no length change was observed and, therefore, no bending. Starting from 1.5~bar pressure, the SRS starts to replicate all locomotion gaits well. The best replication of serpentine gait is observed at 4~bar-0.25~Hz pressure-frequency combination, as presented in Fig. \ref{fig:RollingSerpentineimage}-a. At 2~bar pressure and 0.25-0.50~Hz low-frequency combinations, the SRS shows considerably slow serpentine patterns. At low frequencies, when pressure increases from 2 to 4~bar, the serpentine pattern improves. However, at higher frequencies, the pressure amplitude distorts the gait pattern replication, as presented in Fig. \ref{fig:MovementTracking}-C. This is due to these fast pressure changes not reaching PMAs in realtime through long pneumatic lines. Therefore the caused propagation delay interferes with replication patterns.

\section{Conclusions \label{sec:ConclusionsFuturework}}

Snakes are highly capable animals with a unique ability to navigate challenging terrains using different locomotion gaits. We proposed a SRS that can replicate various snake locomotion gaits. We derived a full floating base kinematic model. We derived the jointspace trajectories for three gait types, namely serpentine, inward, and outward rolling motion, via an optimization approach.  We experimentally tested the SRS for serpentine and rolling snake locomotion. The SRS successfully tracked the spatial shape trajectories for all the gaits at high-pressure amplitudes and low frequencies as expected. Due to the uniform friction coefficient present in both longitudinal and normal motion directions, no meaningful displacement is observed in SRS for the serpentine gait. In contrast, inward and outward rolling gait successfully achieved locomotion. This is the first time that the rolling locomotion is demonstrated in an SRS.

\bibliographystyle{IEEEtran}
\bibliography{refs}
\end{document}